\documentclass[runningheads]{llncs}

\usepackage[T1]{fontenc}
\usepackage{microtype}
\usepackage{graphicx}
\usepackage{url}
\usepackage[svgnames]{xcolor} 
\usepackage{hyperref}
\usepackage{pdflscape}
\usepackage{tcolorbox}
\tcbuselibrary{skins, breakable}
\usepackage{listings}
\usepackage[labelfont=bf]{caption}
\usepackage[capitalise,nameinlink]{cleveref}
\usepackage{xcolor}
\usepackage{booktabs}   
\usepackage{array}      
\usepackage{amssymb}    
\usepackage{tabularx}
\usepackage{listings}
\usepackage{subcaption}
\usepackage{marginnote}
\usepackage{xcolor}
\usepackage{pgfplots}
\usepackage{subcaption}
\pgfplotsset{compat=1.18}
\definecolor{codegreen}{rgb}{0,0.6,0}
\definecolor{codegray}{rgb}{0.5,0.5,0.5}
\definecolor{codepurple}{rgb}{0.58,0,0.82}
\definecolor{backcolour}{rgb}{0.98,0.98,0.98}
\crefname{section}{Sect.}{Sect.}
\Crefname{section}{Section}{Sections}

\usepackage{placeins} 

\hypersetup{
	pdftoolbar=true,        
	pdfmenubar=true,        
	pdffitwindow=false,     
	pdfstartview={FitH},    
	pdfsubject={},   
	pdfcreator={},   
	pdfproducer={}, 
	pdfkeywords={}, 
	pdfnewwindow=true,      
	colorlinks=true,       
	linkcolor=Brown,          
	citecolor=Brown,        
	filecolor=Brown,      
	urlcolor=Brown           
}

\lstdefinelanguage{yaml}{
  keywords={true,false,null,y,n},
  keywordstyle=\color{blue},
  basicstyle=\ttfamily\tiny,
  commentstyle=\color{gray},
  stringstyle=\color{red},
  sensitive=false
}

\newcommand{\mypar}[1]{\vspace{1pt}\noindent\textbf{#1.}}
\newcommand{\myparit}[1]{\vspace{1pt}\noindent\textit{#1.}}

\begin{document}

\title{Agent Behavior Mining: Generative AI Agent Governance in Business Processes}
\date{November 2025}
\titlerunning{Agent Behavior Mining}
\authorrunning{Vu et al.}

%
\author{
Hoang Vu\inst{1}\orcidID{0009-0002-1656-7207} 
\and Maximilian Körner\inst{1}\orcidID{0000-0001-9274-3271} 
\and Adrian Rebmann\inst{1}\orcidID{0000-0001-7009-4637} 
\and Gabriel Kevorkian\inst{1}\orcidID{0009-0005-6589-8135}
\and Michael Perscheid\inst{1}\orcidID{0000-0002-2331-6775} 
\and Gregor Berg\inst{1}\orcidID{0009-0005-6237-3747} 
\and Timotheus Kampik\inst{1,2}\orcidID{0000-0002-6458-2252}
}
\institute{SAP, Berlin, Germany \\
\email{\{h.vu, maximilian.koerner, adrian.rebmann, gabriel.kevorkian, michael.perscheid, gregor.berg, timotheus.kampik\}@sap.com}
\and Umeå University, Umeå, Sweden}
%
%

\maketitle

\begin{abstract}
As organizations increasingly deploy generative AI agents to automate business processes, they face a governance dilemma: although these agents can increase operational flexibility, their non-deterministic nature challenges the control and standardization that Business Process Management seeks to enforce. This paper addresses this \emph{invisible autonomy risk} by introducing \emph{Agent Behavior Mining}, a governance capability that enables the application of process mining techniques to render generative AI agent decision-making observable and traceable. We (1) improve the understanding of generative AI agent behavior through an event data model that translates granular agent activities—including reasoning traces, tool usage, and token costs—into standardized process logs; (2) instantiate the data model in a multi-agent order-to-cash implementation, demonstrating how process managers can leverage agent logs to detect policy deviations and quantify operational variability; and (3) evaluate the perceived practical utility of the approach in an exploratory study with 18 industry practitioners. The results indicate that practitioners view behavioral transparency as a prerequisite for trust and consider the ability to examine agent reasoning as an important governance requirement for the next generation of AI-driven business processes.

\end{abstract}

\section{Introduction}
\label{sec:introduction}
As organizations increasingly deploy generative AI (GenAI) agents in business processes, they face an \emph{autonomy-control paradox}~\cite{Mohlmann2017Hands}: granting agents autonomy can increase operational flexibility, yet Business Process Management (BPM) governance depends on control, standardization, and auditable compliance~\cite{rosemann2014six}. This paradox exposes organizations to an \emph{invisible autonomy risk}: unlike deterministic, rule-based task automation, GenAI agents driven by large language models (LLMs) operate through opaque, non-deterministic reasoning. This opacity creates risks, including undetected policy violations and compliance failures, as well as unmanaged resource consumption~\cite{xi2025rise}.

Process owners face a \emph{black box challenge}: they observe outputs but lack visibility into \textit{why} decisions were made and \textit{how} resources were used. Existing BPM governance mechanisms, built for repeatable execution, lack constructs to analyze probabilistic behavior of GenAI agents, rendering governance difficult. 

To address these challenges, we propose \textit{Agent Behavior Mining} (ABM) as a technology-enabled BPM governance capability that operationalizes visibility for organizational control. ABM establishes (1) \emph{observability} of agent decisions, (2) \emph{accountability} for agent outcomes, and (3) \emph{controllability} by enabling targeted interventions such as policy guardrails and prompt updates. This supports the detection of compliance drift, containment of variable costs, and the development of empirical trust needed to scale agent adoption. We make three contributions:
\begin{enumerate}
    \item \textbf{Event Data Model for Agent Behavior:} We introduce an event data model that provides the basis for turning agent activities---including reasoning traces and token costs---into standardized event logs, extending \emph{eXtensible Event Stream} (XES) standards for GenAI agent governance.
    \item \textbf{Operational Risk Detection:} We instantiate the data model by implementing a multi-agent Order-to-Cash (O2C) scenario, demonstrating how process mining methods (discovery, conformance, variant analysis) quantify compliance deviations, cost anomalies, and behavioral variability.
    \item \textbf{Practical Utility Assessment:} We conduct an exploratory study with 18 industry practitioners assessing ABM’s practical utility and how behavioral transparency influences trust and adoption decisions in critical workflows.
\end{enumerate}

Our findings indicate that ABM is feasible and perceived as potentially useful, with behavioral variability emerging as a core governance concern.
Practitioners expect transparency gains with ABM and view reasoning-level traceability as a prerequisite for trust. Overall, we argue that ABM can provide a practical foundation for moving from pilots to managed, accountable agent deployments.

The remainder of this paper is structured as follows. \autoref{sec:background} reviews related work. \autoref{sec:motivation} introduces a multi-agent system  (MAS) scenario to illustrate the invisible autonomy risk. \autoref{sec:datamodel} derives requirements and introduces the ABM event data model. \autoref{sec:agentmining} instantiates the model, applies process mining techniques, and validates its utility through a practitioner survey. Finally,~\autoref{sec:discussion} discusses implications and limitations, followed by the conclusion in~\autoref{sec:conclusion}.

\section{Background}
\label{sec:background}
Our work relates to governance in BPM, algorithmic accountability and transparency, process mining, and tracing GenAI agents.

\mypar{Governance in BPM}
Governance in BPM provides the structural mechanisms, decision rights, accountability frameworks, and performance metrics necessary to align process execution with organizational strategy. 
This aligns with the six core elements of BPM, which position governance as a distinct capability alongside strategic alignment and people~\cite{rosemann2014six}. Seminal work by Weill and Ross~\cite{weill2004governance} defines IT governance as the specification of decision rights to encourage desirable behavior. In traditional BPM, this governance relies on bounded, deterministic workflows where deviations are treated as manageable exceptions. However, the deployment of GenAI agents creates an \textit{autonomy-control paradox}~\cite{Mohlmann2017Hands}: organizations want agents to be flexible, yet this flexibility undermines the standardization that governance requires. 
Unlike earlier forms of task automation, such as Robotic Process Automation~\cite{van2018robotic} or classical MAS~\cite{DBLP:journals/aai/JenningsNOO00}, which followed strict rules and protocols, GenAI agents operate through probabilistic reasoning. This introduces risks of unobservable compliance drift that traditional governance mechanisms struggle to detect.

\mypar{Algorithmic Accountability and Transparency}
As algorithms take over operational tasks, they create what Kellogg et al. call a new \emph{contested terrain of control}~\cite{kellogg2020algorithms}, widening an \emph{accountability gap}~\cite{raji2020closing} where the inability to monitor run-time execution prevents responsible AI operationalization. In the context of business processes, algorithmic accountability requires that an automated system provides a justifiable, auditable record of its decision-making---not just of 
\emph{what} it decided, but of \emph{why}. As AI systems become embedded in core business operations, this form of transparency 
becomes a prerequisite for effective governance. Prior work on human trust in AI shows that opaque decision logic can undermine trust, as process owners cannot readily understand why a system acted as it did~\cite{glikson2020human}.
Recent regulatory frameworks, such as the EU AI Act or the NIST AI Risk Management Framework,
codify the need for explainability and auditability in high-risk AI systems. However, these frameworks often remain high-level principles without specifying operational mechanisms for business processes. 
We argue that ABM addresses this gap by repositioning agent governance from a strategic liability into an organizational capability, enabling the behavioral transparency necessary for accountable GenAI agent deployment.

\mypar{Process Mining}
Process mining offers a data-driven approach for process analysis by reconstructing execution flows from event logs, revealing deviations between designed processes and execution reality~\cite{DBLP:books/sp/Aalst16}. Standardized event data models like XES~\cite{gunther2014xes} provide the technical foundation but treat resources as passive executors.
Recent advances in \emph{Object-Centric Event Data} (OCED)~\cite{fahland2024towards} address complex, many-to-many interactions between business objects (e.g., orders, items, shipments) by storing objects and their attributes separately and linking them to events. However, while OCED captures the \emph{structural} reality of interconnected processes, it also does not consider agents as first-class citizens.
To address this, \emph{Agent System Mining} (ASM)~\cite{DBLP:journals/access/TourPK21} 
centers the analysis on the agents themselves, advancing the field by discovering 
agent systems~\cite{DBLP:conf/bpm/TourPKS23} and formalizing their interactions 
through \emph{Agent System Event Data} (ASED)~\cite{DBLP:conf/er/ShenPLK24}.
Extending the XES meta-model, ASED introduces a separate entity for agents, which captures the roles agents play and the organizations to which they belong~\cite{DBLP:conf/er/ShenPLK24}.

Neither OCED nor ASED addresses the unique characteristics of GenAI agents, such as reasoning capabilities or token-based costs. More importantly, because both approaches extend the XES meta-model, they require adapted or entirely new process mining algorithms to leverage their data.
In contrast, our event data model strictly adheres to the XES standard. By using tailored event types and attributes rather than altering the meta-model, our approach remains fully compatible with off-the-shelf process mining tools, significantly lowering the entry barrier for GenAI agent governance.
Crucially, our proposed model does not conflict with existing frameworks. It can integrate with ASED, which allows agent metadata (e.g., roles) to be stored in separate entities and linked to events, and with OCED to explicitly model the objects agents interact with. In both cases, users can continue to leverage the GenAI-specific elements introduced in our framework.

\mypar{Tracing GenAI Agents}
Emerging \emph{AgentOps} platforms, such as LangSmith, LangFuse, and Phoenix, already provide token-level cost tracking, latency monitoring, error detection, and reasoning traceability for LLM-based agents~\cite{dong2024agentops}, often with OpenTelemetry compatibility~\cite{opentelemetry2024genai}. However, a critical gap remains: designed primarily for machine learning and DevOps engineers, these platforms surface individual execution traces rather than structurally formatted event data~\cite{moshkovich2025beyond}. Because they lack standard event log semantics, where low-level executions must be categorized into standardized, non-unique activity types, these traces cannot be readily analyzed using process discovery, conformance checking, or variant analysis.

To bridge this gap, ABM provides the structural foundation required to perform process mining directly on these agent traces. It structures the event data to capture essential information for process improvement on \emph{accountability}, \emph{decision causality}, \emph{operational performance}, and \emph{financial impact}. In contrast with other approaches, ABM treats \emph{tool usage}, \emph{LLM model configuration}, \emph{token cost}, and \emph{reasoning traces} as first-class process attributes. This capability allows organizations to move beyond monitoring \textit{what} agents do, toward greater visibility into \emph{how} they reason, \emph{what} they cost, and \emph{how well} their execution
aligns with business policies.
We argue that process mining can evolve from a retrospective analysis tool into an active enabler of \emph{trustworthy AI}, providing the fact-based visibility that can help operationalize high-level governance principles.

\section{The Invisible Autonomy Risk}
\label{sec:motivation}
\begin{figure}[t]
\centering
\includegraphics[width=0.75\linewidth]{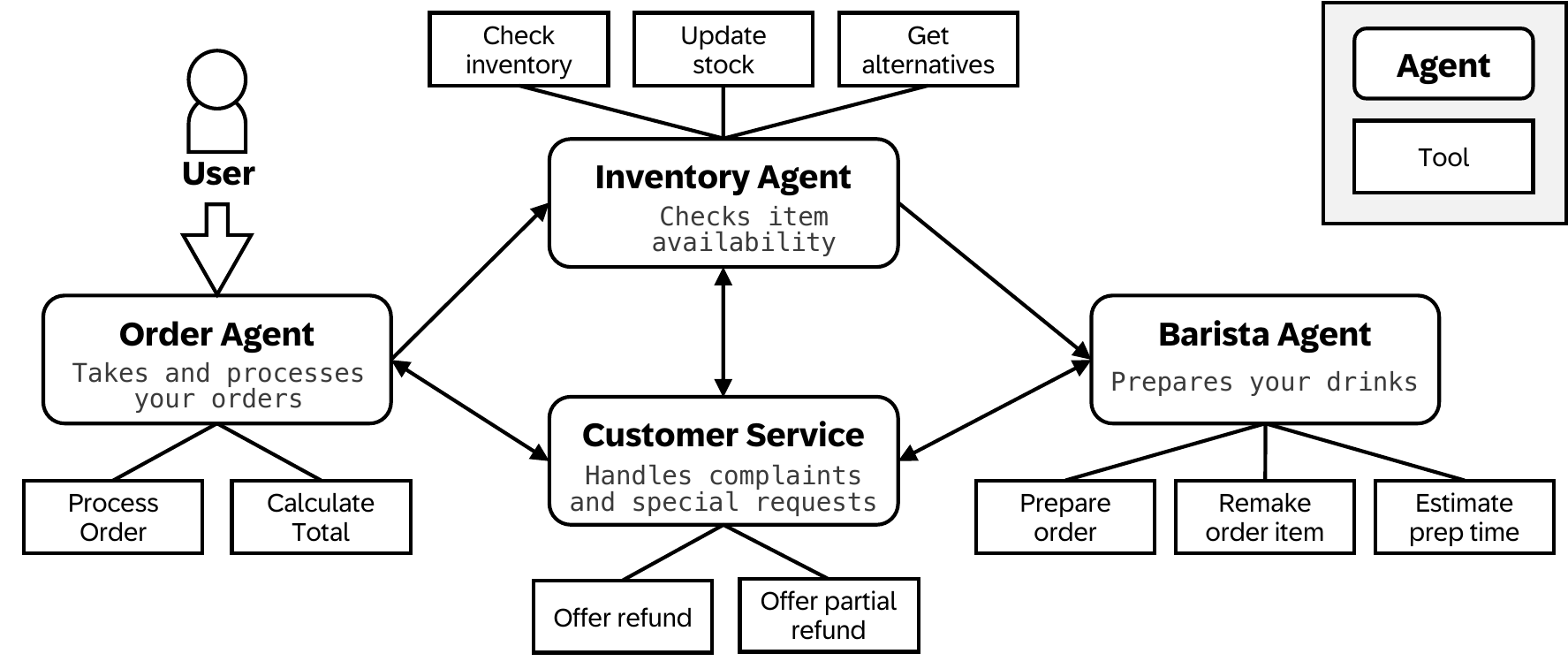}
\caption{Order-to-Cash multi-agent system with four GenAI agents.}
\label{fig:coffee-mas}
\end{figure}

Consider a MAS that manages an O2C process for a coffee shop (\autoref{fig:coffee-mas}). Four autonomous GenAI agents coordinate the work: the \emph{Order Agent} interprets customer intent, the \emph{Inventory Agent} manages stock, the \emph{Barista Agent} executes production, and the \emph{Customer Service Agent} adjudicates refunds. 

Traditional workflows embed business logic explicitly in process models, making decisions traceable to their triggering rules. GenAI agents internalize logic probabilistically, creating a governance blind spot: when the \emph{Order Agent} handles \textit{``large caramel latte, oat milk, refund if delayed,''} it interprets ambiguous terms without exposing its reasoning---leaving managers unable to verify whether a policy was applied or a new rule was hallucinated. This opacity compounds into systemic risk: unnecessary restocking by the \emph{Inventory Agent} becomes an unmonitored financial leak, and adversarial prompts can silently induce the \emph{Customer Service Agent} to grant undeserved refunds---both undetectable without access to reasoning chains.

Existing observability tools do not close this gap: agent frameworks produce verbose logs optimized for debugging, not governance. Without a standardized event data model
translating these traces into process-level data, organizations cannot govern their
digital workforce at scale---leaving them exposed to the very autonomy they sought
to leverage.

\section{Event Data Model for Agent Behavior}
\label{sec:datamodel}
In this section, we present an event data model for GenAI agent behavior by operationalizing the ASED meta-model~\cite{DBLP:conf/er/ShenPLK24} and extending XES~\cite{gunther2014xes} with agent-specific attributes (Contribution 1).
While XES provides the foundation for event representation, it lacks constructs for the specific characteristics of GenAI agents. We therefore derive requirements from operational goals and industry standards (\autoref{sec:requirements}) to instantiate our model (\autoref{sec:components}).

\subsection{Model Requirements}
\label{sec:requirements}
We derive requirements from two perspectives: \emph{goal-oriented}, addressing the
operational objectives of GenAI agent deployment, and \emph{system-driven}, stemming
from agent architectures and observability standards like OpenTelemetry~\cite{opentelemetry2024genai}.

\mypar{Goal-Oriented Requirements}
Adopting the \emph{Devil's Quadrangle}~\cite{dumas2018fundamentals}, we view GenAI
agent deployment as process redesign targeting four performance dimensions: cost, time,
quality, and flexibility.

\myparit{R1 (Cost)}
Unlike traditional software, GenAI agents incur direct per-action costs via token
usage. The model must correlate individual steps with their exact financial impact.

\myparit{R2 (Time)}
Activity durations range from near-instantaneous tool calls to multi-step LLM
reasoning with substantial latency. The model must capture this variance to assess
time-efficiency relative to task complexity.

\myparit{R3 (Quality)}
Non-deterministic execution requires tracing failures to their origin in user intent.
The model must expose the reasoning chain from prompt interpretation to agent decision.

\myparit{R4 (Flexibility)}
Agents autonomously adapt execution paths based on context. The model must capture
distinct sequences to make behavioral variability observable.

\mypar{System-Driven Requirements}
These requirements stem from GenAI agent framework architectures, AgentOps
observability standards, and the structural requirements of process mining platforms.

\myparit{R5 (Behavior)}
The model must capture atomic events---LLM calls, tool executions, and agent
delegation---to isolate logic errors from API or handoff failures and enable
precise root cause analysis.

\myparit{R6 (Semantics)}
Heterogeneous agent frameworks produce traces with varying structures and granularity.
The model must align with OpenTelemetry GenAI semantic
conventions~\cite{opentelemetry2024genai} to ensure expressiveness and prevent
information loss.

\myparit{R7 (Interoperability)}
Proprietary formats create vendor lock-in. The model must produce standardized output
aligned with the XES standard~\cite{gunther2014xes}, accessible to any process mining
platform without costly migration.

\subsection{Event Data Model}
\label{sec:components}
\begin{figure}[t]
    \centering
    \includegraphics[width=0.95\linewidth]{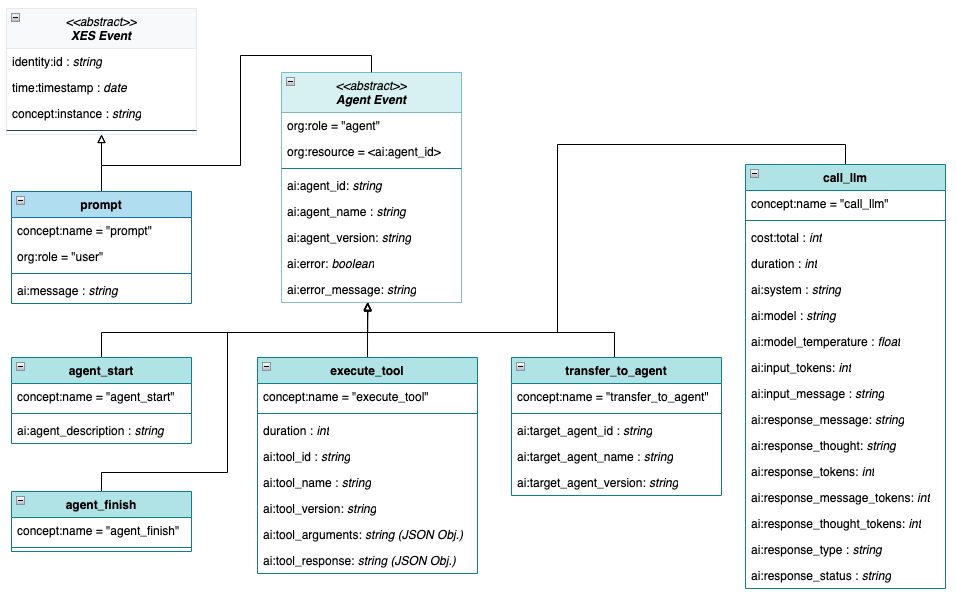}
    \caption{Agent Event Concept Hierarchy}
    \label{fig:event-data-model}
\end{figure}

Based on the above requirements, we propose the event data model visualized in \autoref{fig:event-data-model}. The model extends XES and can complement previous event data models---specifically ASED~\cite{DBLP:conf/er/ShenPLK24} and  OCED~\cite{fahland2024towards}---by adding GenAI-specific event types and attributes that are absent from existing event data models (\autoref{sec:background}).

The event data model is organized as a shallow hierarchy where all events inherit standard attributes from the XES \texttt{concept}, \texttt{org}, and \texttt{time} extensions. The \texttt{prompt} event type captures user inputs within single or multi-turn conversations, establishing the context for subsequent system actions. In multi-turn settings, a case corresponds to a complete conversation session: each user turn is recorded as a \texttt{prompt} event sharing the same \texttt{case\_id}, so the full dialogue context is preserved as a single traceable unit. All remaining event types are agent-centric, derived from an abstract \texttt{AgentEvent} type that factors out common attributes. Specific event types include \texttt{agent\_start} and \texttt{agent\_finish}, which track the overall execution time of agent tasks (R2). Agent operations are captured through \texttt{call\_llm} and \texttt{execute\_tool}, while \texttt{transfer\_to\_agent} represents delegation, a fundamental interaction pattern in MAS (R5). 
Standard attributes are adapted to support business-oriented analysis. \texttt{concept:name} and \texttt{concept:instance} capture the technical event type and a human-readable activity name, respectively (R4); grouping events by these names enhances understandability in process visualizations (e.g., DFGs). To assess costs (R1), various \texttt{*\_tokens} attributes track LLM consumption. For long-running tasks, we introduce a \texttt{duration} attribute to complement the standard \texttt{time:timestamp} (R2). To ensure quality and transparency (R3), internal reasoning is tracked via \texttt{ai:response\_thought}, while \texttt{ai:error*} attributes detail execution failures. Table~\ref{tab:traceability} provides a traceability matrix summarizing how each requirement is satisfied by the attributes in the event data model.
Crucially, most \texttt{ai:*} attributes align with OpenTelemetry GenAI semantic conventions (R6), ensuring broad compatibility and semantic coverage. Although not formally defined as an XES extension, the model builds upon the XES standard~\cite{gunther2014xes} and corresponding event logs are readily usable by process mining tools supporting the standard (R7). By satisfying these requirements (R1--R7), our model provides a practical foundation for capturing the nuances of GenAI agent behavior, enabling the practical application of existing process mining techniques, as demonstrated in the following section.

\begin{table}[b]
\centering
\caption{Requirement--attribute traceability}
\label{tab:traceability}
\begin{tabular}{lll}
\toprule
\textbf{Req.} & \textbf{Dim.} & \textbf{Satisfying Attributes / Event Types}\\
\midrule
R1 & Cost     & \texttt{cost:total}, \texttt{ai:*\_tokens} (\texttt{call\_llm}) \\
R2 & Time     & \texttt{duration}, \texttt{time:timestamp} \\
R3 & Quality  & \texttt{ai:\{error, error\_message, response\_thought\}} \\
R4 & Flexib.  & \texttt{concept:name}, \texttt{concept:instance} \\
R5 & Behavior & \texttt{agent\_\{start,finish\}}, \texttt{call\_llm}, \texttt{execute\_tool}, \texttt{transfer\_to\_agent} \\
R6 & Semant.  & \texttt{ai:*} (OpenTelemetry-aligned~\cite{opentelemetry2024genai}) \\
R7 & Interop. & XES \texttt{concept}, \texttt{org}, \texttt{time}~\cite{gunther2014xes} \\
\bottomrule
\end{tabular}

\end{table}

\begin{figure}[t]
\centering
\vspace{-4mm}
\begin{minipage}[t]{0.49\linewidth}
\begin{lstlisting}[
    language=yaml,
    basicstyle=\ttfamily\fontsize{7.5}{8.5}\selectfont,
    basewidth=0.48em,
    aboveskip=0pt, belowskip=0pt,
    lineskip=-1.5pt,
    breaklines=true,
    escapechar=|,
    literate={-}{-}1
]
- case_id: "2cd03d3f-..."
  events:
  - concept_name: "prompt"
    time: "2025-11-03T09:28:38.479"
    ai_message: "Can I please have
      an espresso for Max"
  - concept_name: "agent_start"
    time: "2025-11-03T09:28:38.482"
    ai_agent_name: "Order Agent"
    concept_instance: "[Order Agent] starts"
\end{lstlisting}
\end{minipage}
\hfill
\begin{minipage}[t]{0.49\linewidth}
\begin{lstlisting}[
    language=yaml,
    basicstyle=\ttfamily\fontsize{7.5}{8.5}\selectfont,
    basewidth=0.48em,
    aboveskip=0pt, belowskip=0pt,
    lineskip=-1.5pt,
    breaklines=true,
    escapechar=|,
    literate={-}{-}1
]
  - concept_name: "call_llm"
    time: "2025-11-03T09:28:38.483"
    duration: 691ms
    ai_model: "gpt-4.1-2025-04-14"
    ai_input_tokens: 372
    ai_response_message_tokens: 27
  - concept_name: "execute_tool"
    time: "2025-11-03T09:28:39.178"
    duration: 3.7s
    ai_tool_name: "process_order"
    ai_tool_args: '{"order":[...],"customer":"Max"}'
    ...
\end{lstlisting}
\end{minipage}
\vspace{-3mm}
\caption{Agent event log fragment.}
\label{fig:event-log-sample-yaml}
\vspace{-4mm}
\end{figure}

\autoref{fig:event-log-sample-yaml} illustrates the model through a
concrete O2C execution trace. The \texttt{prompt} event establishes
case context via \texttt{ai\_message} (``Can I please have an espresso
for Max''), while \texttt{agent\_start} attributes subsequent actions
to the ``Order Agent'' via \texttt{ai\_agent\_name}. The
\texttt{call\_llm} event captures model identity (\texttt{gpt-4.1}),
token counts (372 input, 27 output), and execution duration;
\texttt{execute\_tool} logs the action payload via \texttt{ai\_tool\_args}
(\texttt{\{"order":[...],...\}}) to enable decision reconstruction.
Together, these attributes allow organizations to distinguish cost
sources, trace reasoning back to user intent, and analyze operational
efficiency across execution variants.

\section{Agent Behavior Mining in Practice}
\label{sec:agentmining}
This section demonstrates how ABM can address governance challenges in GenAI agents. 
We implemented the multi-agent O2C scenario\footnote{Repository contains the multi-agent O2C implementation, event log with 371 cases, survey instruments, and evaluation dashboards. \url{https://github.com/A-rebmann/ABM/}} to generate event data based on our proposed event data model; the repository includes the implementation, generated event logs, survey instruments, and evaluation dashboards to support conceptual reproducibility. Subsequently, we conducted a two-phase study where we first applied ABM to surface governance-supporting insights (Contribution 2). We then validated the perceived utility of these insights in a study with 18 industry practitioners (Contribution 3).

\subsection{Study Design}
\label{agentmining_study}
We followed Design Science Research~\cite{hevner2010design} and 
conducted an exploratory study to assess artifact utility in a realistic organizational context. We pursued \emph{analytical generalization}~\cite{yin2009case}, using the findings to refine theoretical propositions about GenAI agent governance rather than to estimate population-level effects. We recruited 18 practitioners via purposive sampling to ensure adequate diversity of roles: \textbf{strategic} (4 Enterprise Architects, 1 Product Owner, 1 IT Manager), \textbf{operational} (6 Consultants, 2 Process Managers, 1 Business Analyst), and \textbf{control} (1 Security Expert, 1 Data Scientist, 1 Developer). Participants’ AI experience ranged from none (2) to beginners (9), intermediate (6) and expert (1), supporting assessment across skill levels.

\mypar{Data Collection Protocol}
We used a four-phase protocol to reduce common method bias~\cite{podsakoff2003common}: (1) \textbf{Context setting} (30~min) with a standardized briefing on invisible autonomy risks and the O2C architecture; (2) \textbf{Simulation} (60~min) in which participants interacted with four GenAI agents (Order, Inventory, Barista, Service), generating 371 execution traces, including induced failures (e.g., hallucinations, loop errors), which we converted to our event data model via a framework-agnostic, extensible approach compatible with major agent frameworks; (3) \textbf{Insight elicitation} (30 min) where participants explored four dashboard views---process discovery, conformance checking, performance analysis, and variant analysis---and were asked to identify governance-relevant observations, including compliance violations, cost anomalies, and behavioral variants; and (4) a \textbf{Feedback phase} (20~min) comprising a structured survey~\cite{fowler2013survey} covering \emph{Perceived Usefulness} (TAM-based~\cite{davis1989perceived}), \emph{Transparency}, and \emph{Adoption Barriers}, with Likert items and open-ended responses analyzed through thematic analysis~\cite{braun2006using}: responses were inductively coded by one researcher, grouped into recurring themes, and reviewed by a second researcher to ensure consistency. Validity was strengthened through anonymized responses, pilot-tested instruments, and immediate post-exercise administration. All agents utilized the \texttt{gpt-4.1} model; implications of this choice are discussed in Section~\ref{limitations_evaluation}.

\subsection{Operational Risk Detection}
\label{agentmining_operational}

Leveraging the 371 execution cases of the O2C coffee shop example, we instantiate our event data model to validate its utility for real-world governance.

\begin{figure}[b]
\centering
\includegraphics[width=0.95\linewidth]{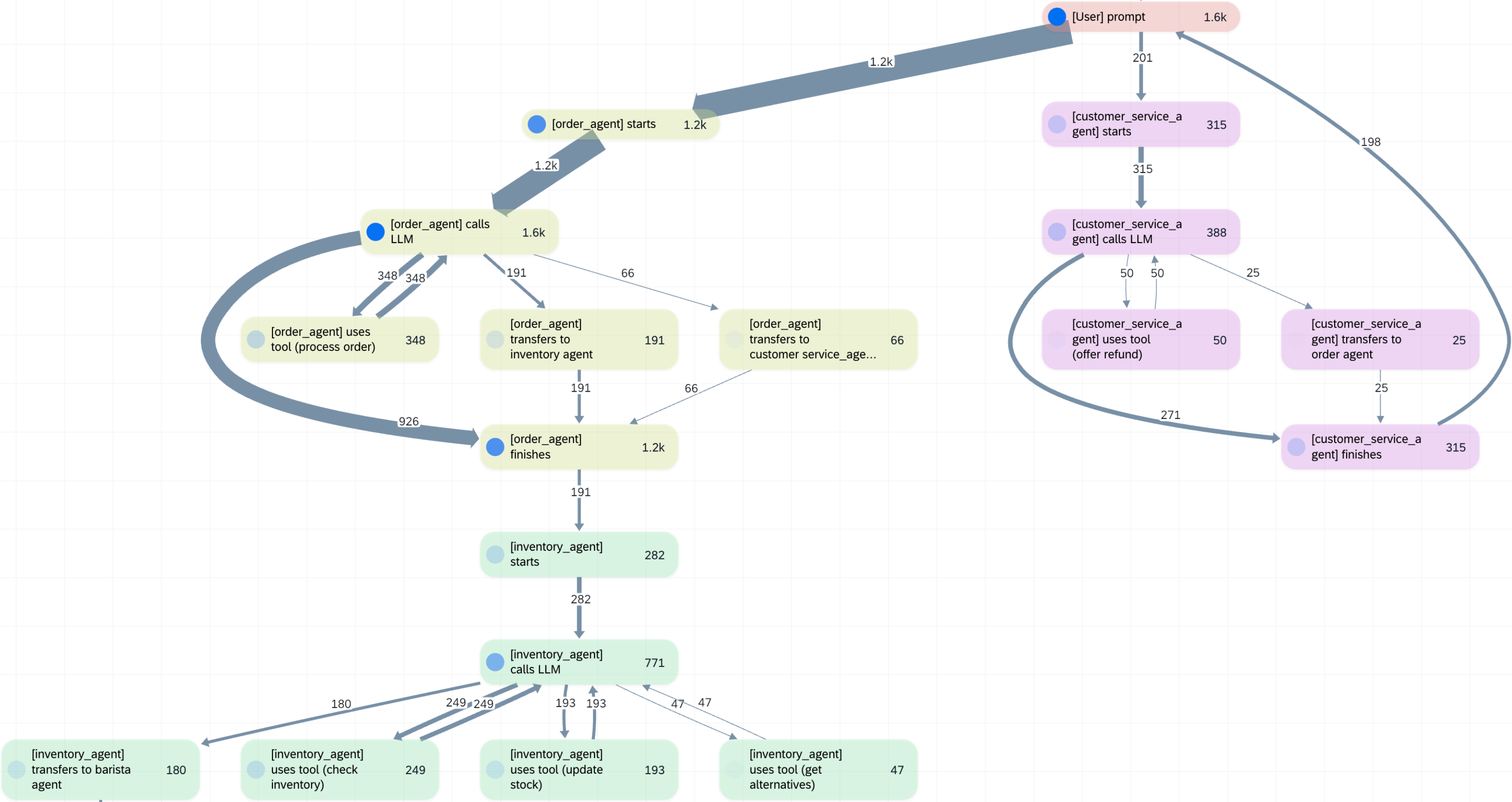}
\caption{Process Discovery: Visibility of Execution Patterns.}
\label{fig:dfg}
\end{figure}

\noindent\textbf{Process Discovery.} 
The directly-follows graph (\autoref{fig:dfg}) captures agent behavior and interactions (R5). The \emph{Order Agent} orchestrates 1.2k instances via \texttt{process order} (348) and delegates to \emph{Inventory Agent} (191) or \emph{Customer Service Agent} (66); \emph{Inventory Agent} executes \texttt{check inventory} (249) before handing off to \emph{Barista Agent} (180). For process owners, this transforms the \emph{black box} of agents' thought processes into a tangible process map.

\begin{figure}[t]
\centering
\includegraphics[width=0.95\linewidth]{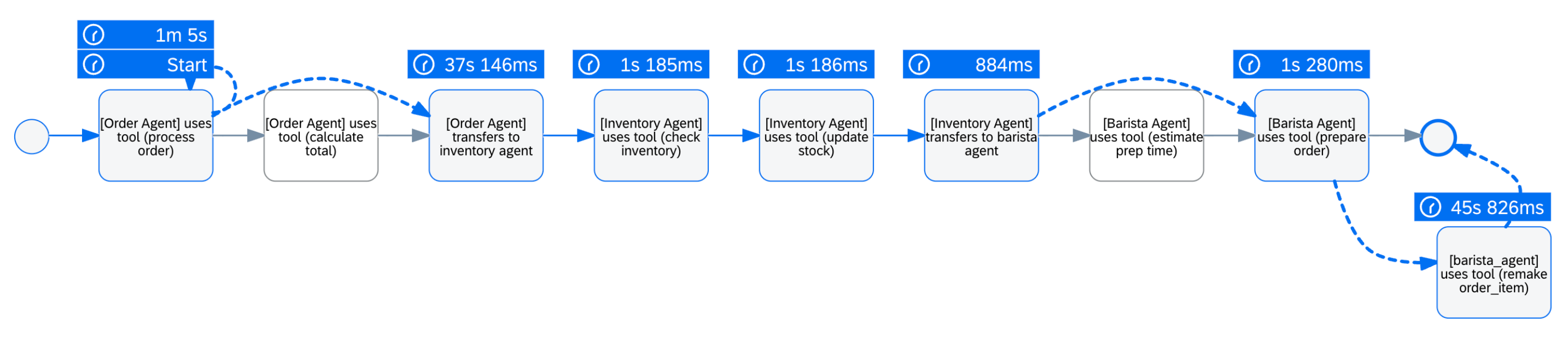}
\caption{Conformance Checking: Specification Alignment Assessment.}
\label{fig:conformance}
\end{figure}

\noindent\textbf{Conformance Checking.} Conformance checking supports failure analysis (R3) by comparing traces against specifications (\autoref{fig:conformance}): \emph{Order Agent} omits \texttt{calculate\_total}; \emph{Barista Agent} skips \texttt{estimate\_prep\_time} and then executes \texttt{remake\_order\_item} (insertion violation). Model alignment with tracing standards (R6) and tool interoperability (R7) enables compliance rate assessment and violation categorization. This quantifies gaps between intended versus actual execution, supporting quality assurance for non-deterministic agents and revealing when agents prioritize customer satisfaction over compliance—providing risk officers with evidence for guardrails and retraining before systemic risks escalate.

\begin{figure}[b]
\centering
\includegraphics[width=0.85\linewidth]{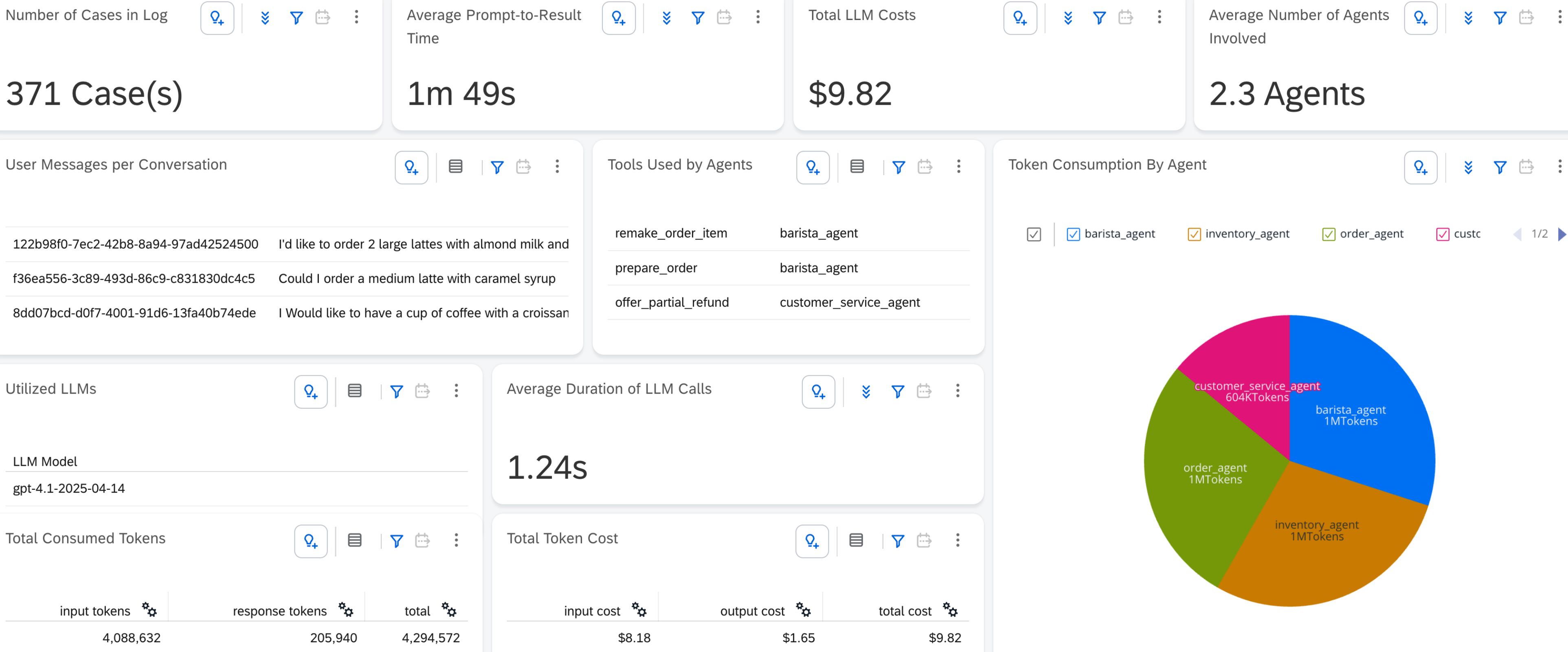}
\caption{Performance Analysis: Quantifiable Metrics.}
\label{fig:performance}
\end{figure}

\noindent\textbf{Performance Analysis.} The performance dashboard (\autoref{fig:performance}) operationalizes cost and time metrics for agent behavior (R1–R2). For 371 cases using \texttt{gpt-4.1-2025-04-14}, we observed 1m 49s average prompt-to-result time, 1.24s average LLM-call latency, 4{,}294{,}572 total tokens (4{,}088{,}632 input, 205{,}940 output) at \$9.82 (\$8.18 input, \$1.65 output), and 2.3 agents per case. Token attribution shows \emph{barista}, \emph{inventory}, and \emph{order} agents each consuming~1M tokens, versus 604k for \emph{customer service}. This enables process managers to benchmark deployments and target costly outliers.

\begin{figure}[t]
\centering
\includegraphics[width=0.82\linewidth]{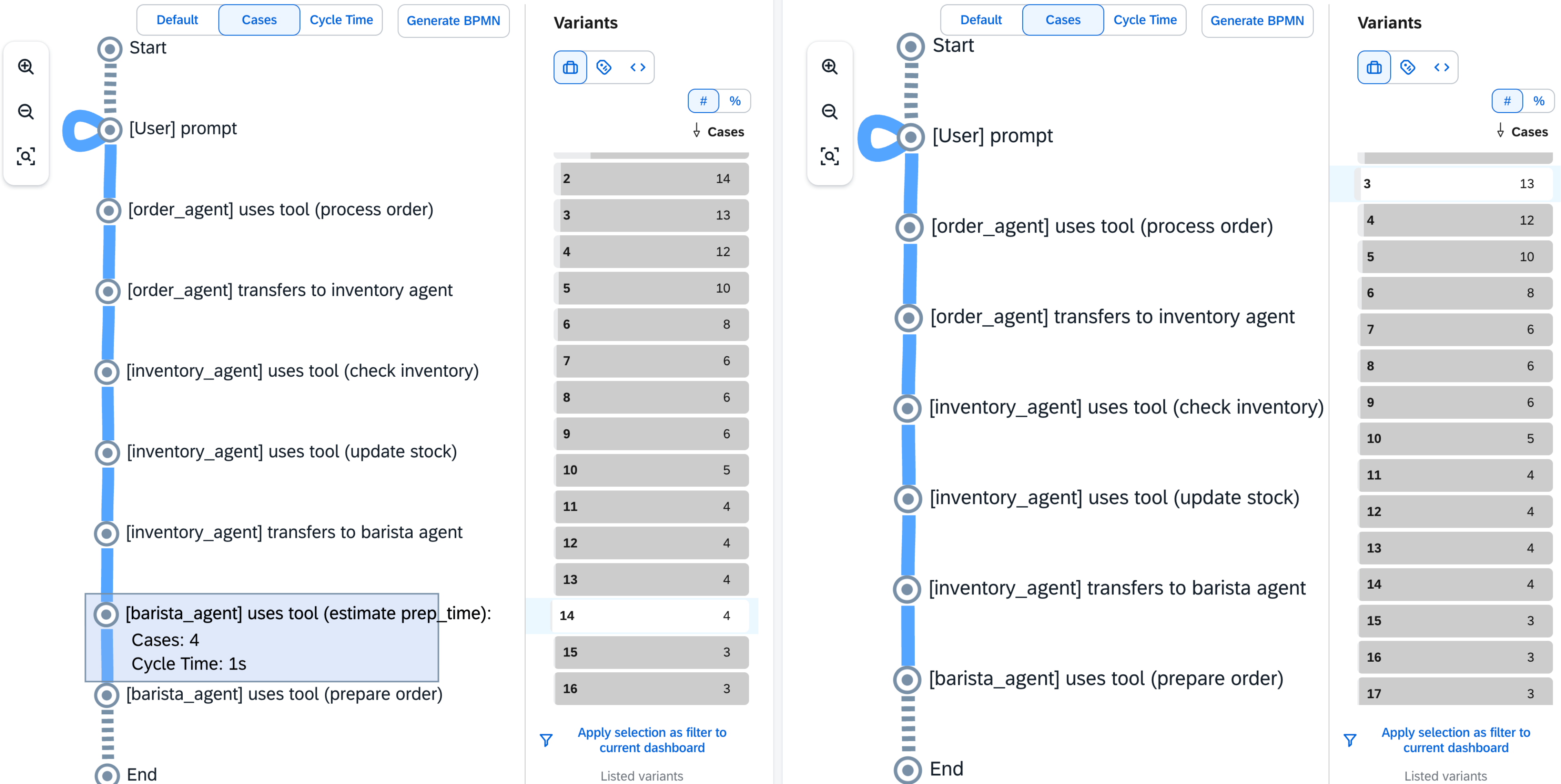}
\caption{Variant Analysis: Observable Execution Differences.}
\label{fig:variant}
\end{figure}

\noindent\textbf{Variant Analysis.} Variant analysis captures GenAI agent variability (R4) (\autoref{fig:variant}). For semantically similar orders, Variant A (13 cases): \emph{Barista Agent} invokes \texttt{estimate\_prep\_time} before \texttt{prepare\_order}; Variant B (4 cases): estimation skipped. Both complete via different reasoning paths. The model enables measuring variant frequency and comparing operational characteristics. We argue that ABM can make behavioral drift observable, potentially enabling organizations to standardize agent behavior and ensure consistent service quality despite probabilistic variance.

\subsection{Practical Utility Assessment}
\label{agentmining_practical}
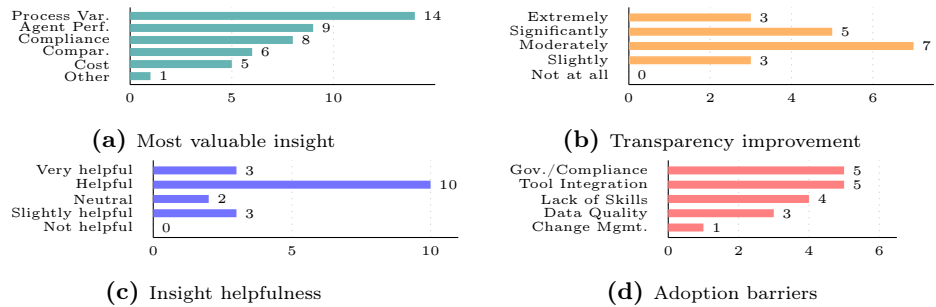
\begin{figure}[b]
\centering
\begin{subfigure}[b]{0.46\linewidth}
  \begin{tikzpicture}
    \begin{axis}[
      xbar, xmin=0, xmax=15,
      xtick={0,5,10},
      width=\linewidth, height=2.6cm,
      bar width=3pt,
      clip=false,
      ymin=0.3, ymax=6.7,
      ytick={1,2,3,4,5,6},
      yticklabels={Other, Cost, Compar., Compliance,
                   Agent Perf., Process Var.},
      tick style={draw=none},
      yticklabel style={font=\tiny},
      xticklabel style={font=\tiny},
      nodes near coords,
      nodes near coords style={font=\tiny, anchor=west},
      axis lines*=left,
      xmajorgrids=true,
      grid style={dotted, gray!40},
    ]
    \addplot[fill=teal!60, draw=none] coordinates {
      (1,1)(5,2)(6,3)(8,4)(9,5)(14,6)
    };
    \end{axis}
  \end{tikzpicture}
  \caption{\scriptsize Most valuable insight}
  \label{fig:survey_valuable}
\end{subfigure}
\hfill
\begin{subfigure}[b]{0.46\linewidth}
  \begin{tikzpicture}
    \begin{axis}[
      xbar, xmin=0, xmax=7.5,
      xtick={0,2,4,6},
      width=\linewidth, height=2.6cm,
      bar width=3pt,
      clip=false,
      ymin=0.3, ymax=5.7,
      ytick={1,2,3,4,5},
      yticklabels={Not at all, Slightly, Moderately,
                   Significantly, Extremely},
      tick style={draw=none},
      yticklabel style={font=\tiny},
      xticklabel style={font=\tiny},
      nodes near coords,
      nodes near coords style={font=\tiny, anchor=west},
      axis lines*=left,
      xmajorgrids=true,
      grid style={dotted, gray!40},
    ]
    \addplot[fill=orange!60, draw=none] coordinates {
      (0,1)(3,2)(7,3)(5,4)(3,5)
    };
    \end{axis}
  \end{tikzpicture}
  \caption{\scriptsize Transparency improvement}
  \label{fig:survey_transparency}
\end{subfigure}

\vspace{0mm}

\begin{subfigure}[b]{0.46\linewidth}
  \begin{tikzpicture}
    \begin{axis}[
      xbar, xmin=0, xmax=11,
      xtick={0,5,10},
      width=\linewidth, height=2.6cm,
      bar width=3pt,
      clip=false,
      ymin=0.3, ymax=5.7,
      ytick={1,2,3,4,5},
      yticklabels={Not helpful, Slightly helpful,
                   Neutral, Helpful, Very helpful},
      tick style={draw=none},
      yticklabel style={font=\tiny},
      xticklabel style={font=\tiny},
      nodes near coords,
      nodes near coords style={font=\tiny, anchor=west},
      axis lines*=left,
      xmajorgrids=true,
      grid style={dotted, gray!40},
    ]
    \addplot[fill=blue!55, draw=none] coordinates {
      (0,1)(3,2)(2,3)(10,4)(3,5)
    };
    \end{axis}
  \end{tikzpicture}
  \caption{\scriptsize Insight helpfulness}
  \label{fig:survey_insights}
\end{subfigure}
\hfill
\begin{subfigure}[b]{0.46\linewidth}
  \begin{tikzpicture}
    \begin{axis}[
      xbar, xmin=0, xmax=6.5,
      xtick={0,2,4,6},
      width=0.82\linewidth, height=2.6cm,
      bar width=3pt,
      clip=false,
      ymin=0.3, ymax=5.7,
      ytick={1,2,3,4,5},
      yticklabels={Change Mgmt., Data Quality,
                   Lack of Skills, Tool Integration,
                   Gov./Compliance},
      tick style={draw=none},
      yticklabel style={font=\tiny},
      xticklabel style={font=\tiny},
      nodes near coords,
      nodes near coords style={font=\tiny, anchor=west},
      axis lines*=left,
      xmajorgrids=true,
      grid style={dotted, gray!40},
    ]
    \addplot[fill=red!50, draw=none] coordinates {
      (1,1)(3,2)(4,3)(5,4)(5,5)
    };
    \end{axis}
  \end{tikzpicture}
  \caption{\scriptsize Adoption barriers}
  \label{fig:survey_barriers}
\end{subfigure}

\caption{Practitioner evaluation results (n=18): perceived value,
transparency improvement, insight helpfulness, and adoption barriers.}
\label{fig:practitioner-eval}
\end{figure}

We complemented the quantitative log analysis with practitioner feedback  to assess the artifact's perceived value for decision-making (\autoref{fig:practitioner-eval}). Note that this phase measures TAM-based perceived usefulness rather than demonstrated governance outcomes: no objective measures of task accuracy or decision quality were collected, and participants reviewed dashboards under laboratory conditions without making real governance decisions.

\mypar{Primacy of Variance Control}
Behavioral inconsistency emerged as the primary governance concern: 78\% of respondents selected \emph{Process Variants} as the most valuable insight, exceeding \emph{Agent Performance} (50\%) and \emph{Compliance} (44\%) (\autoref{fig:survey_valuable}). This pattern held across roles (Process Managers to Security Experts), suggesting that, for these practitioners, visualizing and controlling non-deterministic drift may matter more than traditional speed or cost metrics in probabilistic systems.

\mypar{Transparency and Trust}
ABM appeared to make opaque agent behavior more visible to participants: 83\% reported ``Moderate'' to ``Extreme'' transparency improvement (\autoref{fig:survey_transparency}). This perception was consistent across experience levels; a Security Expert noted that granular logs were essential for \emph{``seeing the stability of the process technically end-to-end,''} indicating that ABM can bridge trust barriers for risk-averse stakeholders.

\mypar{Decision Support}
72\% rated the insights ``Helpful'' or ``Very Helpful'' (\autoref{fig:survey_insights}). Higher prior GenAI agent experience was associated with greater confidence in using ABM for governance,
suggesting that novices may benefit from immediate transparency while experienced practitioners may be better able to translate granular traces into control actions. An Enterprise Architect requested \emph{``a generated paragraph explaining... why the final result appeared,''} validating the capture of reasoning traces while motivating summarization layers in future iterations.

\mypar{Adoption Barriers}
Despite high utility, participants identified barriers to production deployment (\autoref{fig:survey_barriers}): \emph{Governance/Compliance} (25\%) and \emph{Tool Integration} (25\%) dominated, followed by \emph{Lack of Skills} (20\%). This distribution suggests a governance triad for adoption: alignment with compliance frameworks, seamless integration into established toolchains, and capability building. Qualitative feedback from a Consultant reinforced this, stressing that \emph{``automating the integration of agent traces into [standard tools] will be key for adoption,''} implying that embedding agent governance into existing BPM suites may be preferable to standalone operational silos.


\section{Discussion}
\label{sec:discussion}
As organizations move from GenAI agent pilots to productive deployments in their business processes, governance shifts from technical feasibility to operational control. This section examines the implications of ABM, outlines considerations for BPM research, and discusses validity threats and limitations.

\subsection{Implications for AI Governance}
Based on the observability and transparency gains evidenced above, we identify the following areas where ABM could support the governance of GenAI agents as an organizational capability, though empirical validation in production settings remains a direction for future work.

\mypar{Operationalized Compliance}
Conformance checking detected non-deterministic failures, such as the \emph{Order Agent} omitting \texttt{calculate\_total} (\autoref{fig:conformance}). This suggests ABM can facilitate the verification of rules and regulations
through traceable events, potentially enabling compliance officers to pinpoint specific execution traces where policy guardrails were breached.

\mypar{Strategic Resource Allocation}
Performance analysis revealed token consumption disparities---a \emph{Barista Agent} consuming nearly 1M tokens versus a \emph{Customer Service Agent}'s 604k (\autoref{fig:performance}). This granular attribution may enable targeted placement of \emph{human-in-the-loop} controls at high-cost or high-risk points while leaving routine cases automated.

\mypar{Data-Driven Continuous Improvement}
By visualizing execution from user prompt to agent output (\autoref{fig:dfg}), ABM can enable an integration of GenAI agents into established improvement cycles. This can support a more systematic and disciplined approach to prompt engineering~\cite{schmidt2024towards}, shifting from ad-hoc changes toward metric-guided refinement.

\mypar{Trust and Change Management}
83\% of practitioners reported improved transparency (\autoref{fig:survey_transparency}), with security experts valuing the ability to inspect process stability. 
Prior research on trust in AI highlights transparency as an important antecedent to trust~\cite{glikson2020human}. ABM dashboards may help stakeholders evaluate agent behavior and support change management in risk-averse settings, potentially supporting more collaborative human-agent working arrangements.

\mypar{Organizational Readiness}
Adoption barriers clustered evenly: tool integration (25\%), governance/compliance (25\%), and skills (20\%) (\autoref{fig:survey_barriers}). This suggests that scaling ABM may require cross-functional teams blending process analysis with AI engineering, integration into existing BPM suites to avoid silos, and data minimization policies to address Personally Identifiable Information (PII) risks.

\subsection{Implications for the BPM Discipline}
Building on Dumas et al.’s agenda for explainable and adaptable AI-augmented process management~\cite{dumas2023ai}, we argue that ABM suggests several disciplinary adaptations for extending process mining to autonomous agents.

\mypar{Resources as Probabilistic Agents}
Traditional BPM treats resources as executors of control flows~\cite{dumas2018fundamentals}. GenAI agents embed decision logic within resources, blurring the model-executor distinction. Building on the agenda outlined above~\cite{dumas2023ai}, this suggests BPM should complement control-flow modeling with explicit agent goals and constraints.

\mypar{Prompts as Process Artifacts}
Process logic has traditionally resided in BPMN models and business rules~\cite{dumas2018fundamentals}. In agentic processes, prompts increasingly function as de-facto process definitions, shifting the focus of behavioral governance away from explicit models. This motivates treating prompts as governed artifacts---specified, versioned, and managed with rigor comparable to process models---while ABM links prompt configurations to execution outcomes.

\mypar{From Efficiency to Safety}
BPM optimization has historically prioritized time, cost, and quality~\cite{dumas2018fundamentals}. Autonomous agents introduce risks (hallucinations, opaque reasoning) that elevate safety as a first-class
governance concern. ABM provides event-level data to detect unsafe patterns and assess whether behavior remains within acceptable boundaries.

\mypar{Variability as Intrinsic}
BPM literature views variability as deviation to minimize~\cite{reijers2021business}. For GenAI agents, behavioral variability is intrinsic---agents legitimately follow different reasoning paths for similar
cases. ABM can make variants explicit, enabling distinction between adaptive problem-solving and non-compliant drift, and suggests the value of developing normative theories of \emph{acceptable variance} for probabilistic executors.

\mypar{From Episodic Redesign to Continuous Adaptation}
Canonical BPM lifecycles assume episodic design-execution-redesign progression~\cite{van2007supporting}. Agent behavior evolves with model or prompt updates even when process models remain stable. This suggests the value of lifecycle models with tighter feedback loops between mining insights and
governed interventions.

\subsection{Threats to Evaluation Validity}
\label{limitations_evaluation}
While the preceding implications highlight ABM's broad potential, our empirical findings must be interpreted with caution due to the exploratory nature and specific setup of our case study.

\mypar{Construct Validity}
We measured utility through perceived helpfulness. While perception precedes adoption~\cite{davis1989perceived}, it does not equate to objective outcomes; participants may have rated potential value rather than immediate efficacy. The insight elicitation used guided dashboard exploration rather than structured tasks with measurable outcomes, and no objective measures of decision accuracy or response quality were collected, which limits the strength of this assessment.

\mypar{Internal Validity}
The workshop setting may have introduced novelty or desirability bias. We mitigated this through realistic scenarios, genuine failure modes, and anonymized responses; however, the short duration of the study limited the assessment of long-term learning curves.

\mypar{LLM Choice} All evaluated agents ran on \texttt{gpt-4.1}. Cost, latency, behavioral variability, and reasoning trace availability are heavily model-dependent: models with stronger instruction-following may exhibit fewer conformance violations, while those without extended thinking yield no \texttt{ai:response\_thought} data. The specific governance insights derived in this study may therefore differ across model families or versions.

\mypar{External Validity}
The 371 execution cases were generated during the workshop under controlled conditions with induced failures, rather than drawn from a production environment; the trace distribution may not reflect the frequency or variety of real-world agent behavior. Our evaluation relies on a single researcher-constructed multi-agent scenario and a limited sample of 18 practitioners, precluding statistical generalization. Furthermore, our findings reflect sequential ReAct agents~\cite{Yao_ReAct}; alternative architectures may present distinct governance challenges. Replication across domains with different process characteristics, such as document-intensive, long-running, or regulatory-heavy processes, is needed to assess broader generalizability. Thus, our current results demonstrate analytical feasibility rather than universal applicability.

\subsection{Limitations of the Proposed Framework}
\label{limitations_framework}
The proposed framework and its data model present several inherent limitations that point toward future research directions.

{\mypar{Reasoning Availability} ABM's failure analysis capability (R3) depends on the availability of \texttt{ai:response\_thought}. However, some LLMs do not perform internal reasoning, and for models that do generate chain-of-thought content, such as OpenAI o-series or Anthropic Claude, availability may further depend on invocation parameters.
Where reasoning summaries are absent, analysts can detect \emph{that} an agent deviated but not \emph{why}, as the agent’s internal rationale remains hidden. In these cases, fields such as \texttt{ai:error\_message}, \texttt{ai:tool\_arguments}, and \texttt{ai:response\_message} may help infer the underlying causes.}

\mypar{Event Granularity} The model captures events at the framework interaction level (\texttt{call\_llm}, \texttt{execute\_tool}, \texttt{transfer\_to\_agent}) but treats intra-LLM processing as atomic. Potential sub-steps, such as multi-step reasoning, within a single \texttt{call\_llm} invocation remain unobservable as discrete events. Extracting finer-grained events from \texttt{ai:response\_thought} content is a promising direction but requires NLP techniques and risks noise from non-deterministic generation.

\mypar{Session-Level Case Notion} While each case can capture a multi-turn conversation, our framework assumes that all messages relate to the same instance or intent established by the user's first message, i.e., we assume that the user does not start a new topic or distinct request within the same conversation, but that all exchanged messages remain related to the initial prompt. Should topics change mid-session, the framework currently cannot dynamically split or join cases based on this shift. Furthermore, concurrent or asynchronous interactions may require richer case correlation beyond a shared session identifier.

\mypar{Data Model Constraints} We based our event data model on XES, as it is formally standardized and widely supported in process mining tools. However, we acknowledge that once standardized, the OCED model may offer advantages for complex scenarios involving numerous concurrent agents operating on diverse, interacting business objects, which we aim to investigate in the future.

\section{Conclusion}
\label{sec:conclusion}
Generative AI agents promise a fundamental shift in operational efficiency, yet they introduce a critical risk: the invisible autonomy of non-deterministic decision-making. As organizations increasingly delegate core business tasks to autonomous systems, the inability to observe and audit why an agent acted becomes a strategic vulnerability. This paper bridges the gap between technical execution and organizational control by introducing Agent Behavior Mining (ABM), a governance capability that transforms opaque agent behavior into auditable, manageable process data.

We have contributed to the management of AI-driven processes by providing: (1) an event data model that captures GenAI agent behavior, enabling rigorous monitoring; (2) an implemented demonstration for detecting operational risks, such as policy deviations and costly variance; and (3) preliminary empirical evidence that ABM can enhance the transparency required for trusted adoption.

While our study represents an initial step within specific architectural boundaries, it charts a clear path for future work. Future work should
(1) integrate ABM event logs with object-centric process mining~\cite{fahland2024towards} to reconstruct end-to-end case choreographies spanning human tasks and agent decisions in hybrid workflows;
(2) design closed-loop governance mechanisms that trigger prompt updates or guardrail adjustments when ABM detects recurring conformance violations above a defined threshold;
(3) conduct longitudinal field studies in real-world organizational deployments to quantify the return on investment of ABM and establish the break-even point for human-in-the-loop oversight;
(4) develop methods to extract structured information from \texttt{ai:response\_thought} content to enable reasoning-level conformance checking;
(5) consider concurrent or asynchronous agent interactions, e.g., when multiple agents are operating in parallel. Also, while ABM is domain-agnostic,
(6) applying it across process domains will establish boundary conditions for its governance capabilities.

Ultimately, ABM aims to empower organizations to move beyond blind reliance on autonomous systems. By establishing a foundation for verifiable governance, we aim to enable enterprises to harness the speed of GenAI without sacrificing the control necessary for sustainable operations.

\medskip
\mypar{Acknowledgments}
\label{sec:acknowledgment}
The authors thank their SAP colleagues from Signavio and Research \& Innovation, especially Thomas Kowark, Martin Sprengel, Ralf Teusner, and Robert Witt for their valuable contributions to this project.

\end{document}